\documentclass[conference]{IEEEtran}
\IEEEoverridecommandlockouts
\usepackage{cite}
\usepackage{amsmath,amssymb,amsfonts}
\usepackage{algorithmic}
\usepackage{graphicx}
\usepackage{textcomp}
\usepackage{xcolor}
\def\BibTeX{{\rm B\kern-.05em{\sc i\kern-.025em b}\kern-.08em
    T\kern-.1667em\lower.7ex\hbox{E}\kern-.125emX}}

\usepackage{booktabs}
\usepackage{orcidlink}
\usepackage{siunitx}
\usepackage[edges]{forest}
\usepackage{subcaption}

\usepackage{pgfplots}
\pgfplotsset{width=7cm,compat=1.8}

\usepackage[acronym,nomain,acronymlists=main]{glossaries}
\glsdisablehyper
\newacronym{fpga}{FPGA}{field programmable gate array}
\newacronym{ann}{ANN}{artificial neural network}
\newacronym{rtl}{RTL}{register transfer logic}
\newacronym{pl}{PL}{programmable logic}
\newacronym{ip}{IP}{intelectual property}
\newacronym{qnn}{QNN}{quantised neural network}
\newacronym{dl}{DL}{deep learning}
\newacronym{dpu}{DPU}{deep learning processing unit}
\newacronym{ps}{PS}{processing unit}
\newacronym{cnv}{CNV}{convolutional neural network}
\newacronym{bnn}{BNN}{bynary neural network}
\newacronym{relu}{ReLU}{rectified linear unit}
\newacronym{mobilenet}{MobileNet}{mobile neural network}
\newacronym{onnx}{ONNX}{open neural network exchange}
\newacronym{ml}{ML}{machine learning}
\newacronym{hls}{HLS}{High Level Synthesis}
\newacronym{simd}{SIMD}{single instruction multiple data}
\newacronym{pe}{PE}{process element}
\newacronym{rgcc}{RG2C}{red grape chunk classification}
\newacronym{rgb}{RGB}{red-green-blue}
\newacronym{fn}{FN}{false negative}
\newacronym{fp}{FP}{false positive}
\newacronym{ai}{AI}{artificial intelligence}
\newacronym{gpu}{GPU}{graphical processing unit}
\newacronym{cpu}{CPU}{computer processing unit}
\newacronym{cnn}{CNN}{convolution neural network}
\newacronym{asic}{ASIC}{application-specific integrated circuit}
\newacronym{tpu}{TPU}{TensorFlow processing unit}
\newacronym{ncs}{NCS}{neural computing stick}

\begin{document}

\title{Red grape detection with accelerated artificial neural networks in the FPGA's programmable logic
\thanks{Identify applicable funding agency here. If none, delete this.}
}

\author{\IEEEauthorblockN{Sandro C. Magalhães}
\IEEEauthorblockA{\textit{INESC TEC}\\
Porto, Portugal \\
{0000-0002-3095-197X}}
\and
\IEEEauthorblockN{Marco Almeida}
\IEEEauthorblockA{\textit{ISEP, PPorto}\\
1201565@isep.ipp.pt\\
Porto, Portugal}
\and
\IEEEauthorblockN{Filipe Neves dos Santos}
\IEEEauthorblockA{\textit{INESC TEC}\\
Porto, Portugal \\
{0000-0002-8486-6113}}
\and
\IEEEauthorblockN{António Paulo Moreira}
\IEEEauthorblockA{\textit{FEUP} and 
\textit{INESC TEC}\\
Porto, Portugal \\
{0000-0001-8573-3147}}
\and
\IEEEauthorblockN{Jorge Dias}
\IEEEauthorblockA{\textit{Khalifa University} \\
Abu Dhabi, UAE \\
{0000-0002-2725-8867}}
}

\maketitle

\begin{abstract}
Robots usually slow down for canning to detect objects while moving. Additionally, the robot's camera is configured with a low framerate to track the velocity of the detection algorithms. This would be constrained while executing tasks and exploring, making robots increase the task execution time. AMD has developed the Vitis-AI framework to deploy detection algorithms into \glspl{fpga}. However, this tool does not fully use the \glspl{fpga}' \gls{pl}. In this work, we use the FINN architecture to deploy three \glspl{ann}, \gls{mobilenet} v1 with 4-bit quantisation, \gls{cnv} with 2-bit quantisation, and \gls{cnv} with 1-bit quantisation (\gls{bnn}), inside a \gls{fpga}'s \gls{pl}. The models were trained on the \gls{rgcc} dataset. This is a self-acquired dataset released in open access. \Gls{mobilenet} v1 performed better, reaching a success rate of \qty{98}{\textbf{\percent}} and a speed of inference of \qty{6611}{\textbf{FPS}}. In this work, we proved that we can use \glspl{fpga} to speed up \glspl{ann} and make them suitable for attention mechanisms.
\end{abstract}


\section{Introduction}

The advent of \gls{ai} is posing significant computing challenges. These algorithms demand high computing power, usually delivered to a power-consuming \glspl{gpu}. This aspect poses significant challenges while computing in mobile robots, when they cannot access external services. 

Heterogeneous computing is the integration of diverse processor systems to address a specific scientific computing challenge. Such platforms consist of a variety of computational units and technologies, including multi-core \glspl{cpu}, \glspl{gpu}, and \glspl{fpga}. These components offer flexibility and adaptability for a broad spectrum of application domains \cite{Andrade2018}. These computational units can significantly boost the overall system efficiency and decrease power consumption by parallelising tasks that demand extensive \gls{cpu} resources over prolonged periods.

Accelerators, such as \glspl{gpu} and \glspl{fpga}, are designed for massively parallel processing. Namely, \glspl{gpu} are designed for conducting matrix multiplications. However, \glspl{cnn} are inherently parallel but unsuitable for matrix representation. \Glspl{asic} are customised designs of \glspl{fpga}, aimed at promising and specifying operation executions. When designed for processing \glspl{cnn}, \glspl{asic} can match the speed of \glspl{fpga}. 

\Glspl{cnn} are the most common type of \glspl{ann} for visual recognition and identification. Several researchers have explored efficient types of \glspl{cnn} to address the challenges of detecting objects in near real-time. Notable examples include Tiny-YOLO \cite{Redmon2016,Redmon2018}, YOLACT \cite{Bolya2019} or other innovative solutions \cite{Howard2017,Sandler2018,Liu2016a}. However, in embedded \glspl{gpu}, these \glspl{ann} are not fast enough yet. Other solutions designed for computation in near real-time in embedded devices are \glspl{fpga}, \glspl{tpu}, Intel \glspl{ncs}, and other dedicated \glspl{asic} \cite{Puchtler2020}. In these devices, quantising the \glspl{ann} for efficient computation \cite{Yang2019b} is common. Magalhães, S.C., \textit{et al.} ~\cite{Magalhaes2023} made a deep benchmark on several different devices and concluded the benefits of using \glspl{fpga} for computing \glspl{dl} models. However, they did not consider the full potential of \glspl{cnn} being deployed on the \glspl{fpga}' \gls{pl}.

In this work, we explore the potential of \glspl{fpga}' \gls{pl} to execute \glspl{ann} for image classification. We will consider the development board AMD ZCU104 \gls{fpga} \cite{AMD2025}, and using PyTorch/Brevitas \cite{Pappalardo2023} and FINN \cite{Umuroglu2017,Blott2018}, we deployed two \glspl{cnn} in the \glspl{pl} of the \gls{fpga}. This work aims to find a solution to classify images in real-time. 

The main contributions of this work are the release of an open-access dataset \gls{rgcc} of small tiles (\qtyproduct{32x32}{px}) of grapes' images for classification purposes. This dataset was used to test and deploy a \gls{bnn} \gls{cnv}, a 2-bit \gls{cnv}, and a 4-bit \gls{mobilenet}.

In summary, we claim that: small-size classification models can be used to identify objects into images; the use of the \gls{pl} is suitable for executing classification \gls{ann}; we released a challenging dataset of tiles of images for detection of fruits in images; we can beat the real-time while inferring images using small size \glspl{ann} in \glspl{fpga}; we can use the \gls{fpga}'s \gls{pl} to deploy \glspl{ann} for the attention mechanisms of robotic systems. 
 
\section{Material and Methods}

This work benchmarks a set of small-sized \glspl{ann} models in a tile dataset of images of grapes. So, in this section, we will explore the required material and the procedures for deploying the \glspl{ann} in the \gls{fpga} 's \gls{pl}.

\subsection{\Gls{rgcc} dataset}

Research suggests that small-sized \gls{ann} operating on dedicated hardware can significantly accelerate the inference process. The new dataset \acrfull{rgcc} was compiled to investigate this. The \gls{rgcc} dataset consists of image segments depicting bunches of red grapes in steep-slope vineyards, classified in a binary manner.

These images were captured using a \acrshort{rgb} camera at Quinta do Seixo, Valença do Douro, Portugal, in video format. An operator utilised a smartphone POCO F2 PRO (for details on the characteristics, see \cite{GSMArena2023}) and manually navigated through the vineyards, recording video samples of the canopy with mature grape bunches. All frames were captured as \gls{rgb} images, allowing for some sequential and repetitive images. The dataset includes \num{1198} images, each with a resolution of \qtyproduct{1920x1080}{px}. 

\begin{figure}[!htb]
    \centering
    \includegraphics[width=\linewidth]{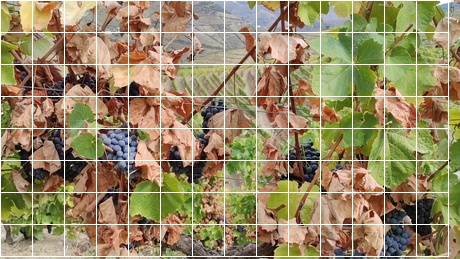}
    \caption{Illustration of the image segmentation scheme for the \gls{rgcc} dataset.}
    \label{fig: rgcc split}
\end{figure}

Subsequently, each image was divided into \qtyproduct{32x32}{px} segments, as illustrated in Figure \ref{fig: rgcc split}. This segmentation process, performed without overlap, yielded a total of \num{10782} images.

Supervised learning techniques were applied. A binary classification method was used, classifying the entire image segment as either containing grapes or not, rather than identifying individual grapes within a bounding box. The dataset was organised in an Image Directory structure. 


The labelling process was conducted in two phases. Initially, a colour threshold technique was used to sort images into their respective class folders (\verb|grape| or \verb|no_grape|), taking advantage of the distinct colour of the red grape bunches against the background. However, some mislabelling occurred, necessitating a manual verification process to correct inaccurately labelled images. An image was classified as belonging to the grape category if it contained at least \qty{25}{\percent} of a grape bunch.

Finally, the dataset was split into three sets considering the acquisition sequence of the images: train (\qty{60}{\percent}), validation (\qty{20}{\percent}), and test (\qty{20}{\percent}). The train set contains  \num{6470} images, the validation set \num{2156} images, and the test set \num{2156} images.

The \gls{rgcc} dataset \cite{Almeida2023} was made publicly available. 

\subsection{FINN and Brevitas}

FINN \cite{Blott2018,Umuroglu2017} is a comprehensive, Docker-based environment provided by AMD. Unlike Vitis-AI \cite{AMDXilinx2022a}, FINN operates directly on the \gls{pl} layer of \glspl{fpga}, converting \glspl{ann} into Verilog code, \gls{rtl} designs, and \gls{ip} blocks. It is a freely available, open-source framework dedicated to constructing, executing, and deploying \gls{qnn} on \gls{fpga} platforms. FINN offers a robust suite of tools and modules that streamline the entire \gls{qnn} development process, from design to deployment, thereby enhancing the efficiency of \gls{dl} models.

\begin{figure}[!htb]
    \centering
    \includegraphics[width=\linewidth]{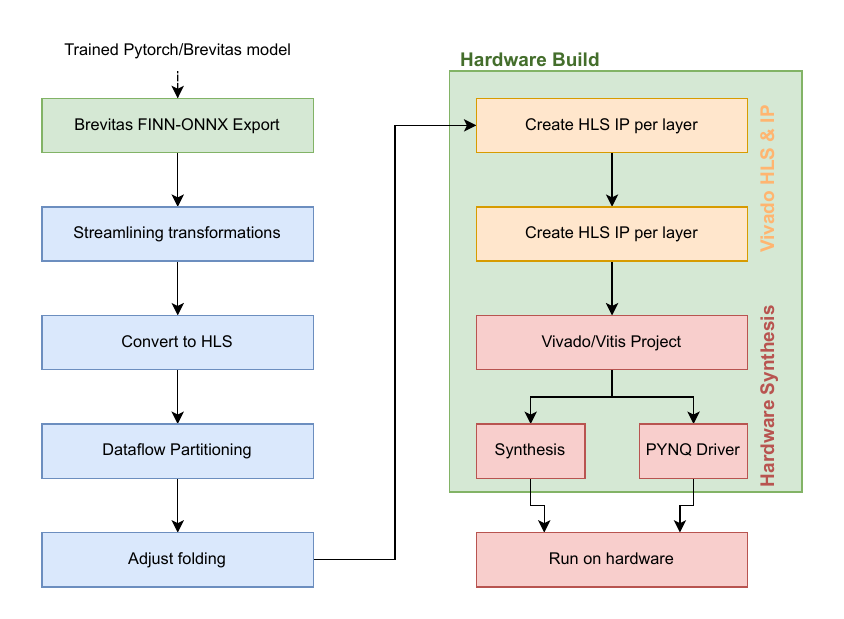}
    \caption{Comprehensive workflow for deploying \gls{dl} models on \glspl{fpga} using FINN.}
    \label{fig: FINN flowchart}
\end{figure}

A workflow for deploying \gls{dl} models onto \glspl{fpga} using FINN is illustrated in Figure \ref{fig: FINN flowchart}.

A notable limitation of FINN is its exclusive compatibility with PyTorch, because of Brevitas \cite{Pappalardo2023} dependency. Consequently, all network designs must be tailored to this specification. Nonetheless, FINN provides the capability to manipulate and adjust \gls{dl} models down to the bit level, thereby enabling the optimisation and compression of network operations to this granularity. This is facilitated by Brevitas, a supplementary library to PyTorch, which redefines common \gls{ann} operations to allow designers to specify the desired bit precision for computations.

In contrast to the conventional approach of employing a \gls{dpu} \gls{ip} core as an intermediary for \gls{dl} models between the \gls{ps} and \gls{pl}, FINN directly deploys a dedicated \gls{ip} core for each model. This strategy is primarily constrained by the finite number of available configurable logic blocks, necessitating careful adjustment of network parallelisation settings and adopting compact-sized \glspl{ann} for implementation.

\subsection{\Glspl{ann} architectures}
\subsubsection{\Gls{cnv}}

\begin{figure}[!htb]
    \centering
    \includegraphics[width=\linewidth]{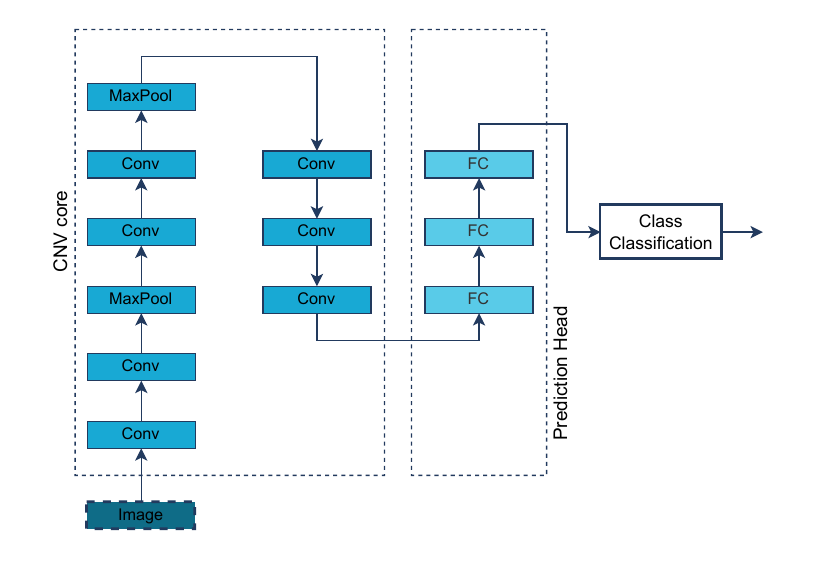}
    \caption{Overview of a simplified diagram of the \gls{cnv}. Conv are convolution layers; MaxPool are maximum pooling layers; and FC are fully connected (or dense) layers.}
    \label{fig: model cnv}
\end{figure}

A \gls{cnv} is structured with sequential layers, including two convolution layers and a maximum pooling layer. Three sequential dense layers power the network's prediction head. The design and structure of this proposed network are depicted in the diagram in Figure \ref{fig: model cnv} and detailed in Table \ref{tab:cnv_model}, both implementing the framework originally introduced by \cite{Umuroglu2017} for the FINN architecture.

To facilitate the deployment of the \gls{cnv} on a \gls{fpga}, we will explore two variants: one with
two-bit operations in each layer and another with single-bit operations per layer. The latter
is recognised as a \gls{bnn} \cite{Qin2020,Yuan2023}. \Glspl{bnn} are designed to achieve optimal efficiency by converting all operations into binary bit-level computations, which are more efficiently processed at the logic level.

\begin{table}[!htb]
\centering
\begin{tabular}{cccc}
\toprule
\textbf{Layer Type}  & {\textbf{Filter Shape}} & {\textbf{Kernel Size}} & {\textbf{Input Size}} \\ \midrule
{Conv}        & \numproduct{64x3x3}                          & \numproduct{3x3}                              & \numproduct{1x3x3x32}                        \\ 
{Conv}        & \numproduct{64x64x3x3}                         & \numproduct{3x3}                              & \numproduct{1x64x30x30}                      \\ 
{MaxPool}     & Pool \numproduct{2x2}                          & \numproduct{2x2}                              & \numproduct{1x64x28x28}                      \\ 
{Conv}        & \numproduct{128x64x3x3}                        & \numproduct{3x3}                              & \numproduct{1x64x14x14}                      \\ 
{Conv}        & \numproduct{128x128x3x3}                       & \numproduct{3x3}                              & \numproduct{1x128x12x12}                     \\ 
{MaxPool}     & Pool \numproduct{2x2}                          & \numproduct{2x2}                              & \numproduct{1x128x10x10}                     \\ 
{Conv}        & \numproduct{256x128x3x3}                       & \numproduct{3x3}                              & \numproduct{1x128x5x5 }                      \\ 
{Conv}        & \numproduct{256x256x3x3}                       & \numproduct{3x3}                              & \numproduct{1x256x3x3}                       \\ 
{Conv}        & --                                 & --                               & \numproduct{1x256x1x1}                       \\ 
{FC}     & \numproduct{256x512}                           & --                                & \numproduct{1x256}                           \\ 
{FC}     & \numproduct{512x512}                           & --                                & \numproduct{1x512}                           \\ 
{FC}     & \numproduct{512x1}                             & --                                & \numproduct{1x512}                           \\ \bottomrule
\end{tabular}
\caption{\Gls{cnv} Architecture}
\label{tab:cnv_model}
\end{table}

\subsubsection{MobileNet v1}

MobileNet v1 \cite{Howard2017} is a \gls{dl} model designed for mobile and low-computing devices. In its original implementation, it is designed for classification tasks. The used version of this classifier was implemented in PyTorch as initially stated. However, some changes in the layers were made to ensure that the \gls{ann} fits in the \gls{fpga} \gls{pl} and is compatible with input data of \numproduct{32x32}~px. At the end of the MobileNet v1 was appended a prediction head composed of an average pooling layer, a fully connected layer and a quantised \gls{relu}.

MobileNet v1, as described by \cite{Howard2017}, is a \gls{dl} model optimised for mobile and low-resource computing devices. Primarily designed for classification tasks, this model's architecture has been adapted for efficient implementation. 

In the version utilised, the \gls{mobilenet} v1 model was implemented using PyTorch, in line with the original description. Modifications were made to its layers to ensure that the \gls{ann} is suitable for deployment on \glspl{fpga}' \gls{pl} and to handle input data with a resolution of \qtyproduct{32x32}{px}. To accommodate the deployment requirements, the architecture of \gls{mobilenet} v1 was appended with a prediction head. This additional component consists of an average pooling layer, a fully connected layer, and a quantised \gls{relu}.


\subsection{Deploying on the \gls{fpga} through FINN}

The process of deploying \glspl{fpga} using FINN is outlined in a sequential manner, as depicted in the referenced FINN flowchart, Figure \ref{fig: FINN flowchart}. The procedure up to the Brevitas export shares similarities with other network deployments and utilises Pytorch. A crucial step involves Brevitas, which is used to implement specific quantisation nodes. These nodes are essential for specifying the bit count for the various operations.

After training, the network is exported to a standard model format known as FINN-ONNX, which is compatible with the \gls{onnx} format (a standard for interoperability of \gls{ml} models between different frameworks \cite{LinuxFoundation2019}). This FINN-ONNX model undergoes several transformations aimed at optimising the network architecture, ensuring it aligns well with AMD-Xilinx \gls{ip} and the \gls{fpga}'s \gls{pl}. Key operations include streamlining \cite{Umuroglu2018}, eliminating unnecessary floating-point operations, merging multiple operations into a single one when feasible, and converting some operations into multiple threshold nodes. These adjustments help reduce the number of operations and enhance parallelisation, resulting in a faster network that effectively utilises the \gls{fpga} space.

Next, the network is transitioned into \gls{hls} layers, followed by conversion into custom \gls{ip} modules or a series of \glspl{ip}. This stage allows for numerous simulations to evaluate the network's optimisation level. The final step involves translating the \glspl{ip} into instructions that the \gls{fpga} can understand.

Control over the network's parallelisation is achieved through two parameters: \gls{simd} and \gls{pe}. \gls{simd} relates to the number of data elements processed concurrently in a single computation, while \gls{pe} refers to the number of parallel computations. The optimisation of these parameters must adhere to \eqref{eq:pe} and \eqref{eq:simd}, to ensure all \gls{fpga} space is efficiently used. In both equations, $H$ and $W$ represent the number of input and output features, respectively. These equations follow the congruence relation notation \cite{Insall2024} and state that the remainder between the integer division of the two values is zero.

\begin{align}
    0 \equiv & H \pmod{\text{PE}}
    \label{eq:pe} \\
    0 \equiv & W \pmod{\text{SIMD}}
    \label{eq:simd}
\end{align}

An advantage of this deployment strategy is the direct implementation of ANNs into logic circuits (configurable logic blocks), optimising performance. Additionally, Brevitas enables bit-level control over the networks. For this analysis, three scenarios were considered: 

\begin{itemize}
    \item A MobileNet v1 with four-bit quantised weights and biases (‘mobilenet\_w4a4’);
    \item A CNV with two-bit quantised weights and biases ('cnv\_w2a2'); and
    \item A CNV with one-bit quantised weights and biases ('cnv\_w1a1').
\end{itemize}

This protocol assures a probable optimisation of the deep neural networks under low-power and high-performing devices. However, they may detect multiple objects simultaneously. The following strategies aim to develop algorithms that can filter the different objects.

\section{Results}

During this section, we explore the ability of the \gls{fpga} to execute \glspl{ann} and identify objects into images. This involves deploying \glspl{ann} as an \gls{fpga} \glspl{ip} through the FINN framework. We present our results to show the capabilities of our method and to support our main claims, which are that the use of the \gls{fpga}'s \gls{pl} to run a small-sized \gls{ann} can be used to identify the objects in classification images from \gls{rgcc}. Secondly, using the \gls{fpga}'s \gls{pl} can outperform the real-time. Thirdly, the use of small-size \glspl{ann} into \glspl{fpga} are suitable for fast inference speed for the attention mechanisms of robotic systems. 

\subsection{Experimental setup}

For the experimental setup, we resorted to the test set from the dataset. After training and deploying the different \glspl{ann}, we tested the results against the test set of the dataset. The experimental process was performed at two different levels. Firstly, we assessed the capability of the \glspl{ann} to classify the different tiles correctly. From this level, we extract performance metrics. Then, we measure the inference speed from the different \glspl{ann} and assess the inference framerate. The second level delivers us the required information to validate the capability of our \glspl{ann} to outperform the real-time, i.e. approximately \qty{25}{FPS}. Both levels aid in validating the suitability of this strategy for applications in the attention mechanisms of robotic perception systems.

\subsection{Models' accuracies to identify objects in images}

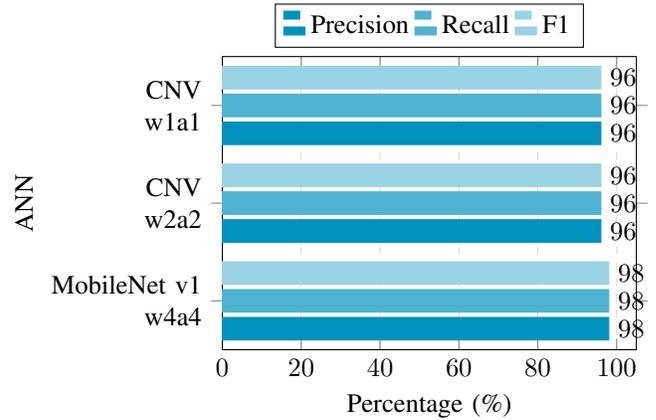
\begin{figure}[!htb]
    \centering

    \begin{tikzpicture}
\begin{axis}[
    width=0.8\linewidth,
    height=5cm,
    xbar,
    xlabel={Percentage (\%)},
    ylabel={ANN},
    ytick={1,2,3},
    yticklabels={
        {MobileNet v1\\w4a4},
        {CNV\\w2a2},
        {CNV\\w1a1}
    },
    yticklabel style={align=right,}, 
    xmin=0,
    xmax=105,
    xtick={0,20,40,60,80,100},
    xmajorgrids=true,
    ymajorgrids=false,
    major grid style={dashed, gray!30},
    legend style={at={(0.5,1.03)}, anchor=south, legend columns=3},
    nodes near coords={\pgfmathprintnumber{\pgfplotspointmeta}},
    nodes near coords align={horizontal},
    bar width=3mm,
    y=13mm,
    enlarge y limits={abs=0.5},
]

\definecolor{maincolor}{HTML}{0092bc}
\definecolor{maincolor70}{HTML}{0092bc}
\definecolor{maincolor40}{HTML}{0092bc}
\colorlet{maincolor70}{maincolor!70}
\colorlet{maincolor40}{maincolor!40}

\addplot[fill=maincolor, draw=maincolor] coordinates {
    (96,3) 
    (96,2) 
    (98,1) 
};

\addplot[fill=maincolor70, draw=maincolor70] coordinates {
    (96,3) 
    (96,2) 
    (98,1) 
};

\addplot[fill=maincolor40, draw=maincolor40] coordinates {
    (96,3) 
    (96,2) 
    (98,1) 
};

\legend{Precision, Recall, F1}

\end{axis}
\end{tikzpicture}
    
    \caption{Inference performance in the evaluation metrics for \gls{fpga} FINN models considering the \gls{rgcc} dataset. The $i,j$ values in $w_ia_j$ report the number of bits being considered for the layers' weights ($w$) and biases (or activations, $a$).}    
    \label{fig: fpga finn results}
\end{figure}

\begin{figure}
    \centering
    \begin{subfigure}[t]{0.49\textwidth}
    \includegraphics[width=\linewidth]{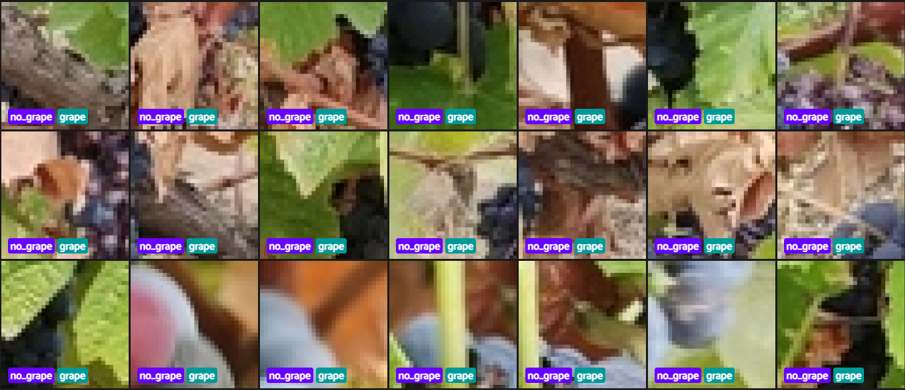}
    \caption{Some false negatives.}
    \label{fig: FINN MobileNet w4a4 FN}
    \end{subfigure}
\hfill
    \begin{subfigure}[t]{0.49\textwidth}
    \includegraphics[width=\linewidth]{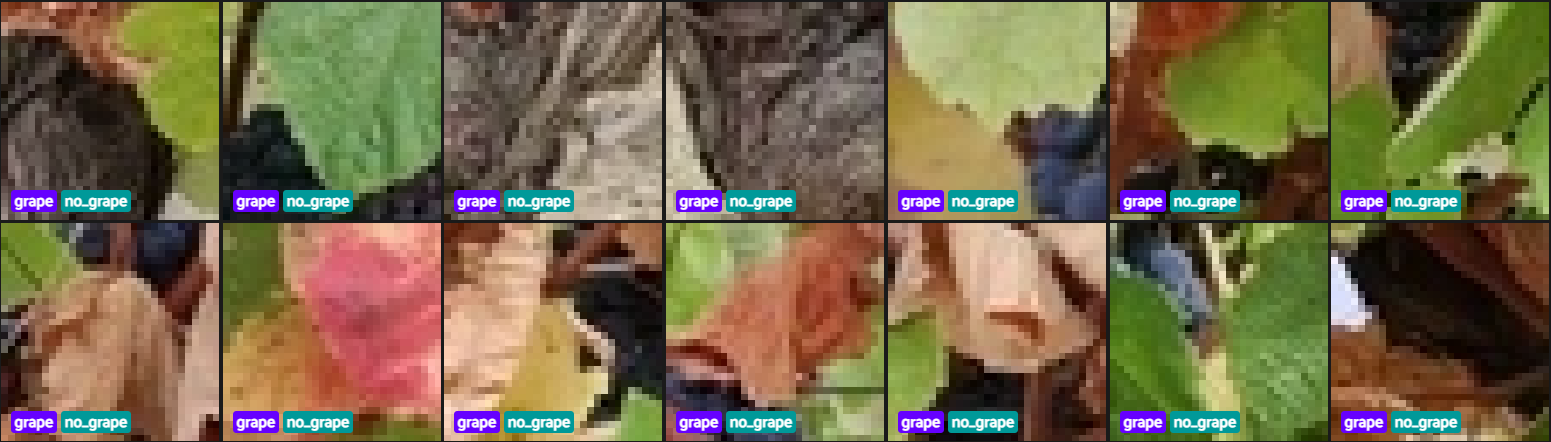}
    \caption{Some false positives.}
    \label{fig: FINN MobileNet w4a4 FP}
    \end{subfigure}
    \caption{FINN MobileNet v1 w4a4.  Cyan labels are the ground truth, and purple labels are the predictions.}
\end{figure}

\begin{figure}[!htb]
    \centering
    \begin{subfigure}[t]{0.49\textwidth}
    \includegraphics[width=\linewidth]{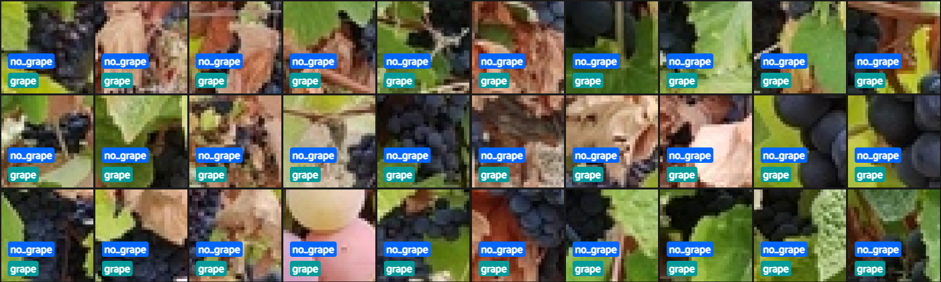}
    \caption{Some false negatives.}
    \label{fig: FINN CNV w2a2 FN}
    \end{subfigure}
\hfill    
    \begin{subfigure}[t]{0.49\textwidth}
    \includegraphics[width=\linewidth]{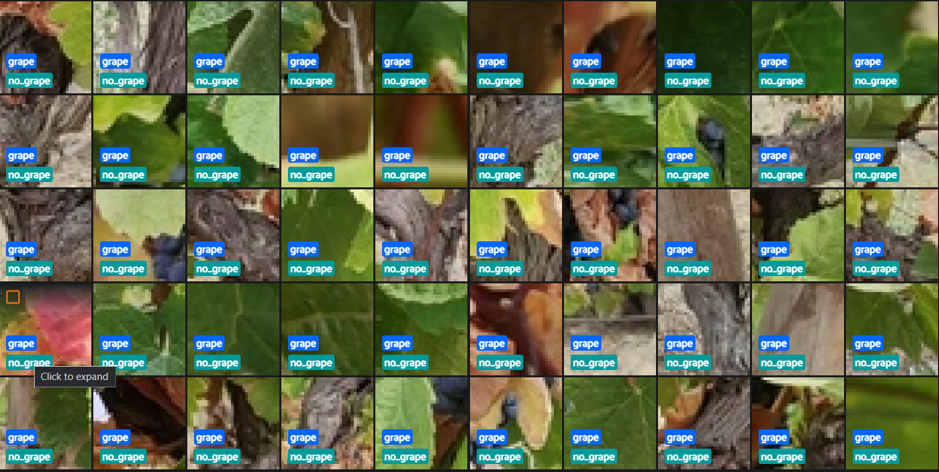}
    \caption{Some false positives.}
    \label{fig: FINN CNV w2a2 FP}
    \end{subfigure}
    \caption{FINN \gls{cnv} w2a2.  Cyan labels are the ground truth, and purple labels are the predictions.}
\end{figure}

\begin{figure}[!htb]
    \centering
    \begin{subfigure}[t]{0.49\textwidth}
    \includegraphics[width=\linewidth]{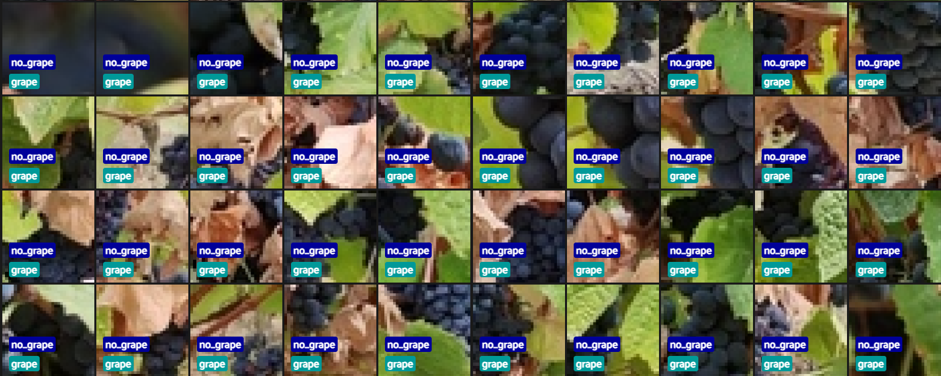}
    \caption{Some false negatives.}
    \label{fig: FINN CNV w1a1 FN}
    \end{subfigure}
\hfill
dataset    \centering
    \begin{subfigure}[t]{0.49\textwidth}
    \includegraphics[width=\linewidth]{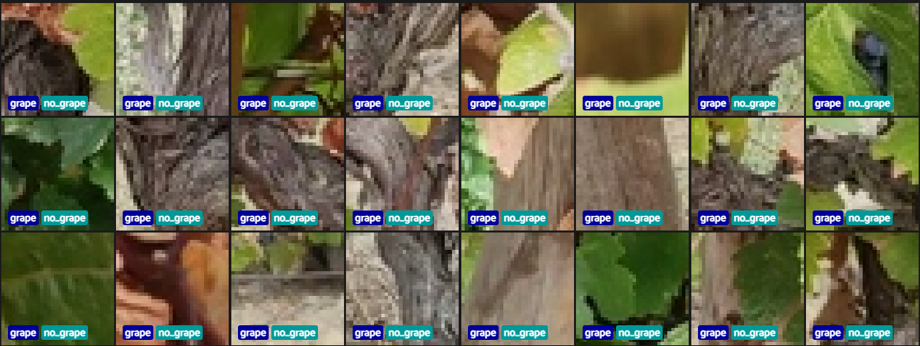}
    \caption{Some false positives.}
    \label{fig: FINN CNV w1a1 FP}
    \end{subfigure}
    \caption{FINN \gls{cnv} w1a1.  Cyan labels are the ground truth, and purple labels are the predictions.}
\end{figure}

FINN is designed to translate \gls{ann} operations into logic levels. However, it's crucial to maintain a reliable metric performance for the classification problem. The performance of various networks deployed in FINN, evaluated using different metrics for the \gls{rgcc} dataset, is documented in Figure \ref{fig: fpga finn results}. Data confirms that the networks deliver consistent and dependable performance. Thus, these networks are suitable for use in robotic applications requiring rapid attention mechanisms. Figures \ref{fig: FINN MobileNet w4a4 FN}, \ref{fig: FINN CNV w2a2 FN}, and \ref{fig: FINN CNV w1a1 FN} showcase some instances of \glspl{fn} for the different models, whereas Figures \ref{fig: FINN MobileNet w4a4 FP}, \ref{fig: FINN CNV w2a2 FP}, and \ref{fig: FINN CNV w1a1 FP} display instances of \glspl{fp}. These examples highlight specific images, specifically partial trunks, that are frequently misidentified as grapes. The high identification accuracies indicate that these systems are suitable for object identification using small-size \glspl{ann}. 


\subsection{Assessing the speed of inference}

\begin{figure}[!htb]
    \centering
    \begin{tikzpicture}
\begin{axis}[
    width=0.8\linewidth,
    height=10cm,
    xbar,
    xlabel={Frames per Second (FPS)},
    ylabel={ANN},
    ytick={1,2,3},
    yticklabels={
        {MobileNet v1\\w4a4},
        {CNV w2a2},
        {CNV w1a1}
    },
    yticklabel style={align=right,font=\small},
    xmin=0,
    xmax=8000,
    xmajorgrids=true,
    ymajorgrids=false,
    major grid style={dashed, gray!30},
    legend style={at={(0.5,1.03)}, anchor=south, legend columns=3},
    nodes near coords={\pgfmathprintnumber{\pgfplotspointmeta}},
    nodes near coords align={horizontal},
    nodes near coords style={font=\small},
    bar width=3mm,
    y=6mm,
    enlarge y limits=0.3,
]

\definecolor{maincolor}{HTML}{0092bc}
\definecolor{maincolor70}{HTML}{0092bc}
\definecolor{maincolor40}{HTML}{0092bc}
\colorlet{maincolor70}{maincolor!70}
\colorlet{maincolor40}{maincolor!40}

\addplot[fill=maincolor, draw=maincolor] coordinates {
    (819.33,3) 
    (819.33,2) 
    (6610.94,1) 
};


\end{axis}
\end{tikzpicture}
    \caption{Processing frame rate for \gls{fpga} FINN models considering the \gls{rgcc} dataset. The $i,j$ values in $w_ia_j$ report the number of bits being considered for the layers' weights ($w$) and biases (or activations, $a$).}
    \label{fig: fpga finn time}
\end{figure}
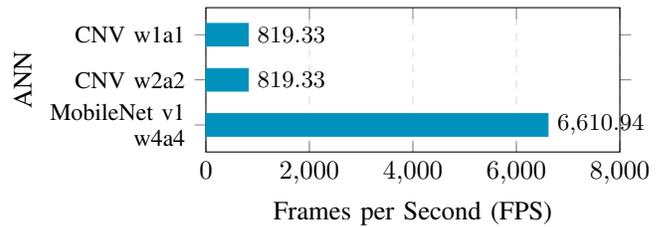

The inference speed was further evaluated, as detailed in \eqref{eq: fps image}. Figure \ref{fig: fpga finn time} illustrates the frame rate for different models deployed on the \gls{fpga}'s \gls{pl} using the FINN framework.  \gls{mobilenet} v1, with four bits for both weights and biases, achieved speeds up to \numproduct{263 x} faster than the \gls{fpga} from \cite{Magalhaes2023} running at \qty{25}{FPS}. Conversely, \gls{cnv} only managed to be \numproduct{32 x} faster. Given that \gls{cnv} is significantly smaller than \gls{mobilenet} v1, there is potential for further optimisation and speed improvements for \gls{cnv} compared to \gls{mobilenet} v1.

Therefore, the use of the \glspl{fpga}' \gls{pl} speeds up the execution of \gls{dl} models in a manner that they become usable for robotics attention mechanisms, acting as an alerting system. At this stage, we can also prove that the use of the \gls{fpga}'s \gls{pl} instead of the \gls{dpu} deliver advantages, and the \glspl{ann} can be sped up to their maximum execution capabilities. 

\begin{align}
    FPS_{chunk} = & \dfrac{1}{t_{avg}} \\
    FPS_{image} = & FPS_{chunk} \times N_c \label{eq: fps image}
\end{align}

\section{Conclusion}

In this article, we explore a strategy to increase the efficiency of executing \glspl{ann} into \glspl{fpga} by optimising the resources available at the \glspl{pl}. We could effectively train and deploy the \glspl{ann} inside the \gls{pl} using FINN. Firstly, we designed a tile-type visual dataset containing images of bunches of grapes in the vineyards. We release the dataset in open access in \cite{Almeida2023}. Then, we trained three different \glspl{ann} models for detecting the fruits in the images. We designed the models using Pytorch and the Brevitas library to convert the various layers into their equivalent for quantised weights. A pipeline from FINN is applied to transform and adjust the \glspl{ann} to fit inside the available \gls{pl}. Through this article, we assess the capability to use this \glspl{ann} deployed on \gls{fpga}'s \gls{pl} to develop an efficient and fast attention mechanism for robotics perception systems. The models could effectively detect the fruits in the images and outperformed the regular speed of cameras, i.e. approximately \qty{25}{FPS}.

\section*{Acknowledgment}


\bibliographystyle{./IEEEtran}
\bibliography{mybibliography}

\end{document}